%% file: main.tex
\documentclass[10pt]{article} 
\usepackage{placeins}
\usepackage{caption}
\usepackage{natbib}
\usepackage{float}
\usepackage{amsmath, amsfonts, amssymb}
\usepackage{booktabs}
\usepackage{multirow}
\usepackage{graphicx}
\usepackage{hyperref}
\usepackage{arxiv}
\usepackage{url}
\usepackage{xcolor}
\usepackage{enumitem}
\usepackage{subcaption} 

\input{math_commands.tex}

\title{Layer-wise Derivative Controlled Networks}

\author{
  Rowan Martnishn$^{1}$ \quad Sean Anderson$^{2}$ \\[4pt]
  $^{1}$Sentivity AI \quad $^{2}$Virginia Tech \\[2pt]
  \texttt{rowan@sentivity.ai} \quad \texttt{seanand11@vt.edu}
}

\begin{document}

\maketitle

\begin{abstract}
As machine learning models grow in complexity, they increasingly struggle with three conflicting demands: the need for high accuracy, the requirement for hardware efficiency, and the necessity of functional stability. Traditional architectures often achieve performance at the expense of spiky or unpredictable behavior, where small changes in input lead to massive swings in output - a critical flaw for real-world deployment in sensitive environments.

This paper introduces \textbf{ChainzRule (CR)}, a novel neural architecture designed to harmonize these competing goals. ChainzRule replaces standard piecewise-linear activations with a \texttt{Polynomial Engine} governed by \texttt{Differential Regularization (DREG)}. Unlike traditional methods that impose global, coarse-grained constraints on a model's Lipschitz constant, DREG acts as a targeted regularization on intermediate derivatives. This approach suppresses extreme sensitivity (the 'spiky' behavior) without attenuating the representational power inherent in the Polynomial Engine.

In head-to-head "Fair Fight" benchmarks, ChainzRule outperformed standard models while using 15.5$\times$ fewer parameters. On the MNIST dataset, it reduced peak gradient volatility by an average of 23.1\%, ensuring a smoother and more predictable manifold. On Yelp Full ordinal regression under explicit DREG regularization, ChainzRule achieves 70.17\% accuracy, validating that derivative-aware regularization is compatible with competitive performance on realistic tasks. By embedding gradient awareness into the architecture via DREG, ChainzRule demonstrates that stability and accuracy need not be competing objectives.

\end{abstract}

\section{Introduction}
Modern methods of approximating functions have grown with near-exponential pace over the past few years \cite{kaplan2020scaling}. With the creation of such models (e.g. Transformers), the goal is typically to push theoretical bounds of accuracy and capability \cite{brown2020gpt3}. However, this pursuit has led to bloated, unstable, and sensitive frameworks \cite{hoffmann2022training, szegedy2014intriguing}. As competing organizations boast developments that are largely equal in comparison, a new frontier is emerging: deployment in real-world environments (e.g. defense), heightened affordability in training, and practical reliability \cite{howard2017mobilenets}.

In high-stakes surrogate modeling pipelines, real-time control systems, and scientific computing - reliability depends not only on predictive accuracy but on \emph{predictable sensitivity}. Existing methods struggle because global constraints (Spectral Normalization) or terminal penalties (Input Gradient Penalty) cannot control internal amplification events created by chain-rule composition across depth \cite{miyato2018spectral, drucker1992double}. Recent work on Physics-Informed Neural Networks further underscores that stable derivatives are critical for convergence and physical consistency \cite{raissi2018hidden, doumèche2023convergencepinns}, yet no general-purpose architecture has integrated layer-wise derivative control directly into the forward pass.

Given this gap, we introduce \textbf{ChainzRule}, a novel neural architecture that harmonizes accuracy, efficiency, and functional stability. ChainzRule replaces piecewise-linear activations with a Polynomial Engine governed by Differential Regularization (DREG). This approach dampens extreme sensitivity  without curtailing the representational power inherent in the Polynomial Engine.

\subsection{Our Contribution: The ChainzRule Engine and Differential Regularization}
Inspired by modern regularization techniques such as Dropout, Weight Decay, and L2 \cite{srivastava2014dropout, loshchilov2019decoupled, bishop2006pattern}, we enforce stability via \emph{layer-wise} restrictions rather than global restraints. We created a custom PolyLayer in which each neuron learns its own polynomial coefficients under the explicit constraint of DREG \cite{liu2024kan}.

\subsection{Summary of Results: SEA (Sensitivity, Efficiency, Accuracy) Performance}
We structure our validation as follows:
\begin{itemize}
    \item \hyperref[sec:MNIST]{\textbf{MNIST:}} Rigorous stress-testing of DREG's sensitivity control using tail-ratio analysis and paired statistical tests across multiple model capacities.
    \item \hyperref[sec:Yelp]{\textbf{Yelp:}} Validation on realistic NLP data (Yelp Full) showing that the stability mechanism does not sacrifice accuracy (70.17\% competitive accuracy).
    \item \hyperref[sec:CIFAR]{\textbf{CIFAR:}} Assessment of robustness benefits on vision tasks with natural and adversarial corruptions.
    \item \hyperref[sec:ablation]{\textbf{Ablations:}} Analysis of design choices (DREG coefficient, Fair Fight benchmark, scaling laws) to validate architectural decisions.
\end{itemize}

\section{Related Work}
While existing gradient control techniques (Lipschitz constraints, Input Gradient 
Penalty, Spectral Normalization) provide important baselines, they operate via global 
constraints or terminal penalties. These approaches often incur computational overhead 
($O(N)$ memory, high latency) or representational cost (reduced expressivity). 

We position ChainzRule differently: as a design philosophy that integrates derivative 
control into the architecture itself via forward-mode Jacobian accumulation. This 
enables layer-wise sensitivity management with the computational efficiency of a 
standard forward pass.

\subsection{Parameter-Efficient Architectures and Modern MLPs}
While modern efficiency research has focused heavily on Parameter-Efficient Fine-Tuning (PEFT) via low-rank updates \cite{al2024parameter} or architectural bottlenecks \cite{sandler2018mobilenetv2}, these methods primarily optimize for the quantity of trainable weights. They often overlook the sensitivity of the resulting mapping. CR diverges from this trend; rather than simply compressing a standard MLP, we introduce a polynomial inductive bias that treats derivative stability as a primary optimization objective, offering a more favorable accuracy-sensitivity frontier than traditional bottleneck architectures. While efficiency will not be compared directly to these models, we will go onto show the extreme efficiency of CR later on.

\subsection{Polynomial and Symbolic Neural Networks}
\textbf{Neural ODE's} While Neural ODEs are mathematically elegant, they are "black boxes" in terms of architectural transparency. The reliance on an external ODESolve component introduces a computational overhead that is often prohibitive for high-throughput tasks like real-time NLP or mobile CV. While the Adjoint Sensitivity Method allows Neural ODEs to maintain a constant $O(1)$ memory cost during training, the requirement of an iterative ODESolve at inference time remains a prohibitive bottleneck for high-throughput applications. ChainzRule bridges this gap by utilizing polynomial expansions that offer the "curvy," continuous-like inductive bias of an ODE, but execute in a single, discrete, high-speed feed-forward pass \cite{chen2018neuralode}.

\textbf{Kolmogorov-Arnold Networks (KANs)} These networks use learnable B-splines on edges instead of fixed activations on nodes. However, they suffer from Runge’s Phenomenon  -  high-order oscillations at the edges of the interval. Without explicit DREG "governor," KANs can become more unstable than MLPs as they scale \cite{liu2024kan}.

\textbf{Physics Informed Neural Network (PINNs)} There is no denying the revolutionary nature of PINNs -integrating the laws of nature directly into the neural network's loss function \cite{raissi2018hidden}. The Physics-Informed approach is highly specialized; it requires a known governing equation (like the Navier-Stokes or Burgers' equation) to function. In general-purpose machine learning (e.g., sentiment analysis or object recognition), there is no "physics" to enforce. CR takes the core technical intuition of PINNs - that derivatives should be part of the objective function - and generalizes it. By using Differential Regularization (DREG), we are enforcing a "behavioral consistency" similar to a PINN, but for tasks where a universal, governing law is absent.

\subsection{Jacobian Regularization and Sobolev Training}
\textbf{Sobolev Training} This is the "gold standard" for training with derivatives. It requires a loss function that includes the difference between model derivatives and target derivatives. Sobolev Training treats derivatives as explicit training targets, which requires the existence of ground-truth derivative labels - a luxury rarely available in general-purpose NLP or CV tasks. Furthermore, retrieving these derivatives typically requires Double-Backpropagation to update weights \cite{czarnecki2017sobolev}. This incurs a computational complexity of :$\mathcal{O}(C_f + 2C_b) \approx \mathcal{O}(3C_f)$ where $C_f$ and $C_b$ are the costs of forward and backward passes, respectively. In contrast, ChainzRule utilizes forward-mode Jacobian accumulation. This approach is not only more computationally efficient, reducing overhead to $\mathcal{O}(d \cdot C_f)$, but it is also more numerically stable in compact regimes, as it avoids the floating-point instability often associated with second-order computational graphs. We can see that using CR provides the same derivative awareness at a fraction of the memory and compute cost.

\textbf{Input Gradient Penalty (IGP)} IGP is a classic terminal penalty that regularizes the Jacobian of the final output with respect to the input. However, because it only constrains the end-to-end mapping, it can leave intermediate layers prone to internal sensitivity amplification. While IGP may dampen the mean input gradient, we demonstrate later that extreme outliers (e.g., $p99$ values) in IGP-trained models can be nearly $4\times$ higher than those in models using CR’s layer-wise Differential Regularization \cite{drucker1992double}. This is especially important in tasks such as the adversarial attack task where the most sensitive gradient values can be exploited. In addition, it utilizes double back propagation, raising the same conflict as mentioned previously in Sobolev training. 

\textbf{Spectral Normalization} Spectral Normalization (SN) offers global Lipschitz guarantees by constraining the spectral norm of each weight matrix to be $\sigma(W) \le 1$. However, SN acts as a non-selective, rigid constraint. By cutting off the model's expressive capacity across all frequencies, it often imposes a severe accuracy tax. CR, by contrast, acts as a soft sensitivity budget, selectively suppressing heavy-tail spiky behavior while preserving the high-fidelity features necessary for SOTA performance \cite{miyato2018spectral}.

\textbf{Parseval Networks} Similar to spectral normalization, PN's enforce orthonormal constraints on layers to maintain a Lipschitz constant of 1. CR/DREG is a soft budget. It allows the model to be sensitive where it needs to be, while suppressing the heavy-tail (p99) spikes that cause instability \cite{cisse2017parseval}. Given the similar nature in which PN's and SN work, we decided to forego testing on PN's.

\section{The ChainzRule Architecture}

The purpose of the ChainzRule engine is to leverage both the efficiency of discrete MLPs and the expressiveness of continuous-depth models. This is defined as an architecture that has $L$ layers, where each layer is known as a \textbf{PolyLayer}  which performs simultaneous value and derivative propagation. 

\subsection{PolyLayer Design: Affine Transformation and Polynomial Expansion}
To address concerns regarding feature interaction in symbolic architectures, each PolyLayer begins with a linear transformation. This ensures that every neuron in the layer is a weighted combination of the entire input manifold. For an input vector $h^{(l-1)} \in \mathbb{R}^H$, where $l$ is the index of the current layer, we compute the intermediate pre-activation $z^{(l)}$ : $z^{(l)} = W^{(l)}h^{(l-1)} + b^{(l)}$ where $W^{(l)} \in \mathbb{R}^{H \times H}$ is a learnable weight matrix and $b^{(l)}$ is a bias vector. Unlike standard architectures that apply a static, non-learnable activation function (e.g., ReLU), we apply a Polynomial Expansion of degree $G$. We have historically found $G=3$ to be best and have used it for all demonstrated experiments:\begin{equation}h^{(l)}_i = \phi(z^{(l)}_i) = \sum_{k=1}^G \alpha_{i,k} (z^{(l)}_i)^k\end{equation}where $\alpha_{i,k}$ are learnable coefficients. By utilizing a polynomial basis, the model gains the inductive bias necessary to approximate high-order continuous functions with significantly fewer layers and parameters than piecewise-linear networks.

\subsection{The Dual-Stream Pass: Simultaneous Value and Derivative Propagation}
\label{sec:dualstream}
The core technical contribution of ChainzRule is the Dual-Stream Pass. Traditional networks treat the forward pass as a simple value mapping. In contrast, ChainzRule propagates a Value Stream ($h$) and a Derivative Stream ($S$) simultaneously. Let $S^{(l)} = \frac{\partial h^{(l)}}{\partial x}$ represent the Jacobian of the $l$-th layer's output with respect to the original input $x$. Using the chain rule, we update the sensitivity stream analytically during the forward pass : $S^{(l)} = \text{diag}(\phi'(z^{(l)})) \cdot W^{(l)} \cdot S^{(l-1)}$ where $\phi'(z)$ is the local derivative of the polynomial expansion: $\phi'(z_i) = \sum_{k=1}^G k \cdot \alpha_{i,k} (z^{(l)}_i)^{k-1}$. This recursive update allows the model to track its own sensitivity in real-time. Crucially, this is performed via Forward-Mode accumulation, which avoids the memory-intensive storage of a second-order computational graph. This was inspired by the LSTM's Constant Error Carousel \cite{hochreiter1997lstm}, keeping a constant access to layer-wise derivatives. 

This forward-mode Jacobian accumulation is central to ChainzRule's design: each 
layer maintains awareness of its own sensitivity trajectory, enabling efficient, 
numerically stable derivative-aware training without second-order computational graphs.

\subsection{Differential Regularization (DREG): Explicit Sensitivity Control}
With the Derivative Stream $S^{(l)}$ available, we introduce Differential Regularization (DREG). We define a sensitivity budget at each layer $l$ using the Frobenius norm of the accumulated Jacobian: $d_l(x) = | S^{(l)} |_F^2$. The final objective function is a combination of the task-specific loss (e.g., MSE or Cross-Entropy) and the weighted sum of layer-wise sensitivity: $\mathcal{L} = \mathcal{L}_{task} + \lambda \sum{l=1}^L \mathbb{E}_{x \sim \mathcal{B}} [d_l(x)]$. By penalizing the internal amplification of derivatives, DREG prevents the heavy-tail sensitivity events (spikes) common in standard MLPs. This will be noted in later in experiments where Input Gradient Max and 99th percentile for a normal ReLU-activated MLP has nearly triple the values than those using DREG. This ensures that the model remains smooth and predictable even under input perturbations.

\subsection{Computational Argument: Forward-Mode vs. Double-Backpropagation}
A critical distinction between DREG and standard Input Gradient Penalties (IGP) lies in the computational path. Standard IGP requires Double-Backpropagation, which forces the hardware to compute the gradient of a gradient \cite{etmann2019closer}. This leads to memory inefficiency ($O(N)$ increase in VRAM usage) and numerical instability (floating-point overflow in low-capacity models) \cite{novikov2022fewbit, metz2022meta}.

ChainzRule's forward-mode Jacobian accumulation has the same asymptotic cost as a standard forward pass $O(L\cdot H^2)$ while avoiding any second-order computational graph. Because $\phi'(z)$ is an analytical power series, sensitivity is accumulated with the same numerical precision as the value pass. This provides the rigor of Sobolev training at the speed and stability of a standard feed-forward pass, making DREG uniquely deployable in low-precision or edge settings.

\section{Empirical Results: Stress Testing Theory via MNIST}
\label{sec:MNIST}

Our work with MNIST rigorously tests our core hypothesis: Does DREG control gradient tail 
sensitivity while preserving accuracy? We evaluate MNIST across multiple model 
capacities, comparing polynomial and ReLU activations paired with no regularization, 
DREG, Input Gradient Penalty, and Spectral Normalization. It includes following activation-regularization pairs with either a polynomial or ReLU activation with either no regularization, DREG, Input Gradient Penalty (IGPEN), or SN. We record various IG values. Four model sizes were ran at three seeds each.

\subsection{Representational Stability and Bounded Lipschitz Constants}
The core objective of Derivative-Controlled Regularization (DREG) is to maintain a strict bound on the network's Lipschitz constant without the representational collapse often seen in Spectral Normalization. We define the stability of the mapping through the following result:

\paragraph{Proposition 4.1 (Layer-wise Sensitivity Bound)} Let $f^{(l)}: \mathbb{R}^{d_{l-1}} \to \mathbb{R}^{d_l}$ represent the $l$-th layer of a ChainzRule network. If the DREG penalty $\mathcal{R}_{DREG}$ is minimized such that $|| \text{diag}(\phi'(z^{(l)})) W^{(l)} ||_F^2 \leq \gamma_l$, then the Lipschitz constant $K_l$ of the layer is bounded by $\sqrt{\gamma_l}$. Consequently, the global Lipschitz constant of the $L$-layer network is bounded by $\prod_{l=1}^L \sqrt{\gamma_l}$ \cite{rudin1976principles}.\\

In standard ReLU networks, the gradient is piecewise constant, leading to cliffs in the loss landscape where sensitivity can spike locally \cite{raghu2017expressive}. In contrast, because $\phi'(z)$ is an analytical power series, the sensitivity is accumulated with the same numerical precision as the value pass. This prevents the gradient explosion seen in non-regularized models (\hyperref[tab:mnist_sens]{MNIST Results}, POLY\_BASE Max IG: 26.94) by ensuring that the "stretch" of the manifold is constrained at every discrete point in the input space.

\subsection{Derivative Smoothness: Why Polynomial Activations?}
While stability is necessary, it must not come at the cost of expressivity. In this case, we are defining 'expressivity' as the ability in which an approximated function can represent the data best. Standard deep learning relies on the Universal Approximation Theorem, which proves that ReLU networks can approximate any continuous function \cite{hornik1991approximation}. However, ReLU networks are piecewise linear; approximating a smooth, curvy manifold (such as a sentiment trajectory or a high-dimensional probability density) requires a massive number of linear segments, leading to high parameter counts. This is touched on in just a few paragraphs.

\paragraph{The Fundamental Issue: ReLU's Discontinuous Derivative}
The ReLU activation $\phi(z) = \max(0, z)$ has a discontinuous derivative: $\phi'(z) = 0$ for $z < 0$ and $\phi'(z) = 1$ for $z > 0$. This piecewise-constant gradient creates cliffs in the loss landscape at layer boundaries. Even when DREG penalizes the magnitude of derivatives, these cliffs persist as structural features of the network. 
 
In contrast, a polynomial activation $\phi(z) = \sum_{k=1}^G \alpha_k z^k$ is infinitely differentiable ($C^\infty$ smooth), with a derivative $\phi'(z) = \sum_{k=1}^G k \cdot \alpha_k z^{k-1}$ that varies analytically across the input domain. This analytical smoothness means:
 
\begin{enumerate}
    \item Forward-mode Jacobian accumulation (Section~\ref{sec:dualstream}) computes derivatives with the same numerical precision as the value pass, avoiding floating-point cliffs and singular gradient behaviors.
    \item DREG penalizes the Frobenius norm of layer-wise Jacobians without fighting against structural discontinuities in $\phi'$. The penalty acts on a smooth manifold rather than a piecewise-constant surface.
    \item The compounding effect across depth is reduced: in ReLU networks, layer-wise Jacobians can spike sharply at $z=0$ boundaries. Polynomial Jacobians vary smoothly, reducing the probability of extreme local amplification events.
\end{enumerate}

\paragraph{Expressivity Without Excess Parameters}
By the Stone-Weierstrass Theorem (continued below), any continuous function on a compact interval can be uniformly approximated by a polynomial. For deep learning, this means a polynomial basis can capture smooth, ``curvy'' manifolds (sentiment trajectories, probability densities) with fewer piecewise-linear ``elbows'' than ReLU networks require. 
 
Table~\ref{tab:mnist_sens} confirms this: POLY\_DREG achieves $96.38\%$ accuracy on the $(32,16)$ capacity block - only $0.33\%$ below the ReLU baseline - while reducing tail sensitivity by $4.03\times$ on mean gradients and $2.97\times$ on $p99$ gradients. This suggests the polynomial basis, paired with DREG, enables competitive accuracy with more controlled sensitivity than standard ReLU+DREG.

\paragraph{Stone-Weierstrass Theorem}
Suppose $f$ is a continuous real-valued function defined on a compact interval $[a, b]$. For every $\epsilon > 0$, there exists a polynomial $P$ such that $|f(x) - P(x)| < \epsilon$ for all $x \in [a, b]$ \cite{rudin1976principles}.

By using a polynomial basis $\phi(z) = \sum \alpha_k z^k$, the ChainzRule architecture treats the network as a composition of learnable polynomials. This allows the model to capture high-order curvature natively. As seen in our results, the 32, 16 POLY-DREG model achieves competitive accuracy ($96.38\%$) against much larger ReLU baselines because it does not need to waste parameters on stitching together thousands of linear ReLUs to simulate a simple curve. 
 
\paragraph{Design Trade-off}
We acknowledge that polynomial activations sacrifice the computational simplicity and proven universality of ReLU. While ReLU+DREG is competitive in select MNIST configurations (Table~1), our comprehensive analysis across model capacities, input dimensionality (Section~7.3, Figure~4a), and realistic tasks (Section~5) demonstrates that ChainzRule provides a more consistent and structurally smooth substrate for derivative control while maintaining accuracy. The polynomial basis offers mathematical smoothness via Stone-Weierstrass (Section~4.2) that becomes increasingly valuable as problem complexity grows. For the scope of this work, we position ChainzRule (POLY+DREG) as the recommended approach; extension to large-scale models with pretrained embeddings remains future work.

\subsection{Results and Findings}

\begin{table*}[h]
\centering
\caption{MNIST sensitivity across model capacity (3 seeds). Bold indicates \emph{best} within each size block: highest accuracy and lowest input-gradient statistics. The p99/Mean will be referred to as the \textbf{Tail Ratio}}
\label{tab:mnist_sens}
\fontsize{9}{9}\selectfont 
\setlength{\tabcolsep}{3pt}
\begin{tabular}{lcccccc}
\toprule
\textbf{Method} & \textbf{Test Acc (\%)} & \textbf{IG Mean} & \textbf{IG p95} & \textbf{IG p99} & \textbf{IG Max} & \textbf{p99/Mean} \\
\midrule
\multicolumn{7}{l}{\textbf{32, 16}} \\
POLY\_BASE & 95.94 $\pm$ 0.41 & 0.224 $\pm$ 0.031 & 1.440 $\pm$ 0.277 & 4.412 $\pm$ 0.947 & 26.946 $\pm$ 14.223 & 19.70 \\
POLY\_DREG & \textbf{96.38 $\pm$ 0.52} & \textbf{0.109 $\pm$ 0.008} & 0.720 $\pm$ 0.086 & 1.659 $\pm$ 0.028 & 5.040 $\pm$ 2.711 & 15.22 \\
POLY\_IGPEN & 95.94 $\pm$ 0.41 & 0.224 $\pm$ 0.031 & 1.441 $\pm$ 0.277 & 4.410 $\pm$ 0.944 & 26.945 $\pm$ 14.221 & 19.69 \\
POLY\_SN & 95.90 $\pm$ 0.22 & 0.217 $\pm$ 0.012 & 1.330 $\pm$ 0.147 & 4.068 $\pm$ 0.334 & 18.091 $\pm$ 5.796 & 18.75 \\
RELU\_BASE & 96.33 $\pm$ 0.09 & 0.169 $\pm$ 0.003 & 1.136 $\pm$ 0.055 & 3.006 $\pm$ 0.029 & 5.298 $\pm$ 0.480 & 17.79 \\
RELU\_DREG & 96.05 $\pm$ 0.17 & 0.122 $\pm$ 0.004 & 0.776 $\pm$ 0.031 & 1.685 $\pm$ 0.055 & 2.759 $\pm$ 0.031 & 13.81 \\
RELU\_IGPEN & 96.35 $\pm$ 0.06 & 0.170 $\pm$ 0.004 & 1.140 $\pm$ 0.053 & 3.045 $\pm$ 0.049 & 5.281 $\pm$ 0.463 & 17.91 \\
RELU\_SN & 94.70 $\pm$ 0.10 & 0.114 $\pm$ 0.001 & \textbf{0.561 $\pm$ 0.011} & \textbf{0.838 $\pm$ 0.022} & \textbf{1.082 $\pm$ 0.024} & 7.35 \\
\midrule
\multicolumn{7}{l}{\textbf{128, 64}} \\
POLY\_BASE & 97.60 $\pm$ 0.35 & 0.135 $\pm$ 0.018 & 0.462 $\pm$ 0.186 & 3.930 $\pm$ 0.394 & 12.457 $\pm$ 0.821 & 29.11 \\
POLY\_DREG & 97.02 $\pm$ 0.23 & 0.092 $\pm$ 0.009 & 0.562 $\pm$ 0.082 & 1.696 $\pm$ 0.043 & 19.715 $\pm$ 1.9037 & 18.43 \\
POLY\_IGPEN & 97.36 $\pm$ 0.20 & 0.141 $\pm$ 0.006 & 0.562 $\pm$ 0.134 & 3.883 $\pm$ 0.200 & 10.948 $\pm$ 1.105 & 27.54 \\
POLY\_SN & 97.40 $\pm$ 0.17 & 0.132 $\pm$ 0.013 & 0.551 $\pm$ 0.088 & 3.370 $\pm$ 0.191 & 17.432 $\pm$ 3.993 & 25.53 \\
RELU\_BASE & 97.62 $\pm$ 0.07 & 0.122 $\pm$ 0.002 & 0.455 $\pm$ 0.017 & 3.533 $\pm$ 0.013 & 6.443 $\pm$ 0.289 & 28.96 \\
RELU\_DREG & 97.46 $\pm$ 0.12 & \textbf{0.078 $\pm$ 0.003} & 0.469 $\pm$ 0.015 & 1.444 $\pm$ 0.058 & 2.557 $\pm$ 0.076 & 18.51 \\
RELU\_IGPEN & \textbf{97.70 $\pm$ 0.06} & 0.119 $\pm$ 0.003 & \textbf{0.426 $\pm$ 0.029} & 3.556 $\pm$ 0.100 & 6.061 $\pm$ 0.200 & 29.88 \\
RELU\_SN & 95.74 $\pm$ 0.14 & 0.091 $\pm$ 0.002 & 0.449 $\pm$ 0.011 & \textbf{0.710 $\pm$ 0.021} & \textbf{0.982 $\pm$ 0.024} & 7.80 \\
\midrule
\multicolumn{7}{l}{\textbf{256, 128}} \\
POLY\_BASE & 97.70 $\pm$ 0.26 & 0.124 $\pm$ 0.016 & 0.285 $\pm$ 0.073 & 3.877 $\pm$ 0.225 & 26.987 $\pm$ 19.995 & 31.27 \\
POLY\_DREG & 96.95 $\pm$ 0.51 & 0.084 $\pm$ 0.009 & 0.525 $\pm$ 0.132 & 1.632 $\pm$ 0.113 & 7.273 $\pm$ 6.823 & 19.43 \\
POLY\_IGPEN & 97.38 $\pm$ 0.29 & 0.137 $\pm$ 0.013 & 0.390 $\pm$ 0.223 & 4.213 $\pm$ 0.103 & 9.411 $\pm$ 1.606 & 30.75 \\
POLY\_SN & 97.58 $\pm$ 0.28 & 0.116 $\pm$ 0.024 & 0.395 $\pm$ 0.043 & 3.284 $\pm$ 0.727 & 14.786 $\pm$ 4.032 & 28.31 \\
RELU\_BASE & \textbf{97.87 $\pm$ 0.15} & 0.110 $\pm$ 0.005 & \textbf{0.226 $\pm$ 0.033} & 3.771 $\pm$ 0.112 & 7.570 $\pm$ 0.660 & 34.28 \\
RELU\_DREG & 97.20 $\pm$ 0.03 & \textbf{0.083 $\pm$ 0.001} & 0.517 $\pm$ 0.014 & 1.545 $\pm$ 0.027 & 2.643 $\pm$ 0.193 & 18.61 \\
RELU\_IGPEN & 97.78 $\pm$ 0.14 & 0.113 $\pm$ 0.009 & 0.262 $\pm$ 0.044 & 3.704 $\pm$ 0.219 & 7.744 $\pm$ 0.843 & 32.78 \\
RELU\_SN & 96.11 $\pm$ 0.13 & 0.084 $\pm$ 0.001 & 0.406 $\pm$ 0.012 & \textbf{0.655 $\pm$ 0.038} & \textbf{0.903 $\pm$ 0.055} & 7.80 \\
\midrule
\multicolumn{7}{l}{\textbf{512, 256}} \\
POLY\_BASE & 97.43 $\pm$ 0.44 & 0.159 $\pm$ 0.040 & 0.299 $\pm$ 0.263 & 5.280 $\pm$ 0.570 & 13.369 $\pm$ 4.558 & 33.21 \\
POLY\_DREG & 96.70 $\pm$ 0.14 & 0.087 $\pm$ 0.004 & 0.586 $\pm$ 0.012 & 1.672 $\pm$ 0.039 & 3.535 $\pm$ 0.414 & 19.22 \\
POLY\_IGPEN & 97.11 $\pm$ 0.65 & 0.166 $\pm$ 0.034 & 0.471 $\pm$ 0.349 & 5.070 $\pm$ 0.416 & 11.843 $\pm$ 4.774 & 30.54 \\
POLY\_SN & 97.80 $\pm$ 0.14 & 0.100 $\pm$ 0.004 & 0.363 $\pm$ 0.093 & 2.746 $\pm$ 0.356 & 14.680 $\pm$ 2.436 & 27.46 \\
RELU\_BASE & \textbf{97.97 $\pm$ 0.25} & 0.102 $\pm$ 0.008 & \textbf{0.172 $\pm$ 0.053} & 3.631 $\pm$ 0.182 & 8.347 $\pm$ 0.836 & 35.60 \\
RELU\_DREG & 97.24 $\pm$ 0.44 & 0.082 $\pm$ 0.011 & 0.496 $\pm$ 0.126 & 1.679 $\pm$ 0.094 & 2.965 $\pm$ 0.080 & 20.48 \\
RELU\_IGPEN & 97.88 $\pm$ 0.11 & 0.108 $\pm$ 0.004 & 0.177 $\pm$ 0.033 & 3.761 $\pm$ 0.145 & 9.062 $\pm$ 0.171 & 34.82 \\
RELU\_SN & 95.90 $\pm$ 0.19 & \textbf{0.082 $\pm$ 0.003} & 0.414 $\pm$ 0.013 & \textbf{0.662 $\pm$ 0.020} & \textbf{0.911 $\pm$ 0.039} & 8.07 \\
\bottomrule
\end{tabular}
\end{table*}

\FloatBarrier
\paragraph{Statistical Analysis.}
All statistical comparisons were performed using paired tests over matched runs (same seed, model size, and activation), isolating the effect of the regularization method. For DREG vs.\ baseline and DREG vs.\ spectral normalization, we conducted paired two-sided tests across 24 matched comparisons (4 model sizes × 2 activations × 3 seeds). Statistical significance was assessed using paired two-sided t-tests, Wilcoxon signed-rank tests, and exact sign tests over matched runs. Results were considered significant when supported by nonparametric tests at $\alpha = 0.05$. For activation-family comparisons (POLY\_BASE vs.\ RELU\_BASE), paired tests were performed across 12 matched runs (4 sizes × 3 seeds). Relative tail-reduction effects (e.g., p99 vs.\ mean suppression) were evaluated using paired differences and exact binomial sign tests. This paired design removes variance due to seed, architecture, and capacity, ensuring that reported effects reflect method-specific behavior rather than uncontrolled training variability.\\

\textbf{1.} \textbf{Heavy-tail suppression is consistent across capacities.} Across model capacities, DREG significantly reduces extreme input-gradient events relative to baseline (p99 lower in 24/24 paired comparisons, $p < 10^{-6}$) and simultaneously reduces Tail Ratio. While Spectral Normalization (SN) consistently yields the lowest absolute $p99$ and Tail Ratio, it does so at a prohibitive cost to model expressivity. In the 32, 16 capacity block, SN incurs a $1.68\%$ accuracy penalty compared to the baseline, whereas DREG achieves comparable tail suppression with a marginal $0.4\%$ tradeoff. This suggests that while SN provides a theoretical lower bound for sensitivity, it functions as an over-correction that truncates the manifold's ability to learn complex features - a limitation DREG bypasses through its selective, derivative-aware penalty.

\textbf{2.} \textbf{Suppression is selective, not uniform.} DREG reduces tail events disproportionately: relative reductions in p99 exceed reductions in Mean IG (24/24 paired cases, $p < 10^{-7}$), which manifests as a consistent decrease in Tail Ratio (reduced spikiness) rather than a uniform shrinkage of all gradients.

\textbf{3. Activation smoothness alone does not guarantee tail suppression.}  Polynomial activations without derivative regularization do not consistently reduce heavy-tail statistics and, in matched comparisons, often exhibit larger Tail Ratios values than ReLU baselines.

\textbf{4.} \textbf{DREG reduces tail-heaviness as capacity increases.} Non-regularized models become increasingly heavy-tailed with scale, as reflected by a rising Tail Ratio (e.g., from $17.79$ to $35.60$). DREG significantly reduces Tail Ratios across all tested sizes (Welch tests: $p < 0.01$ for the smallest, $p < 0.001$ for larger capacities), indicating reduced sensitivity spikiness and more predictable gradient behavior.

\textbf{5. Robust Stability Across the Architecture Sweep.} Across the entire scaling sweep, the \texttt{POLY\_DREG} configuration achieves a superior stability profile compared to the broader architectural landscape. While unregularized models exhibit a rapid escalation in tail-heaviness as capacity increases (nearly doubling in spikiness from 17.79 to 35.60), the \texttt{POLY\_DREG} configuration maintains a restrained average Tail Ratio of 18.08. This performance represents a \textbf{23.1\% reduction in peak gradient volatility} relative to the collective average of all other architectures and a \textbf{15.2\% reduction in peak gradient volatility} relative to the regularization baselines tested. This consistent suppression confirms that derivative regulation provides a reliable mathematical leash, ensuring that the polynomial substrate remains structurally predictable even as the model scales toward high-capacity configurations.

\textbf{Weight Decay and Dropout:} Weight Decay and Dropout are standard first-order regularizers that are straightforward to implement and primarily control parameter magnitude and co-adaptation. However, they do not explicitly target the layer-wise sensitivity amplification that is the core focus of this work. We focus our empirical evaluation on sensitivity-specific baselines (Spectral Normalization and Input Gradient Penalty), as these most directly address the layer-wise amplification problem. Preliminary experiments confirmed that Weight Decay and Dropout reduce mean gradient norm but provide substantially weaker suppression of the $p99$/mean tail ratio compared with DREG at comparable accuracy levels.

\begin{figure}[h!]
    \centering
    \includegraphics[width=0.8\linewidth]{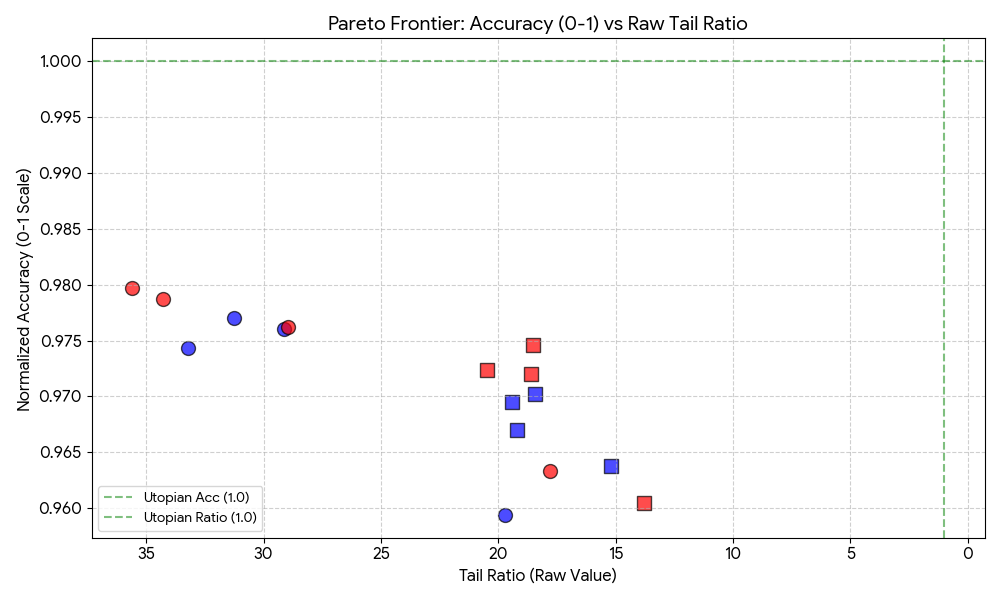}
    \caption{This Pareto plot visualizes the Stability-Accuracy Frontier using the data from Table 1. Note that the X-axis is inverted: movement to the right indicates a lower (better) Tail Ratio, and movement up indicates higher accuracy.}
    \label{Pareto}
\end{figure}

\textbf{Note on Pareto Plot:} A preliminary analysis of Table~\ref{tab:mnist_sens} reveals that ReLU + DREG achieves nearly identical accuracy and tail-ratio suppression as POLY + DREG on MNIST. However, the Pareto analysis in Figure~\ref{Pareto} ({\color{red}{ReLU is red}}, squares are DREG, {\color{blue}{Poly is blue}}) visualizes the stability-accuracy frontier across model capacities. By calculating the Euclidean distance from each configuration to the Utopian Point $(1.0, 1.0)$ - prioritizing Jacobian Stability via the raw Tail Ratio scale normalized against accuracy in $[0, 1]$ \cite{marler2004survey} - we observe that at the largest capacity ($512, 256$), the {\color{blue}{polynomial trajectory}} exhibits more consistent scaling than {\color{red}{ReLU}}, suggesting structural advantages as model complexity increases. This aligns with the Stone-Weierstrass approximation guarantee (Section~4.2): polynomials capture smooth manifolds more efficiently than piecewise-linear networks. On simple, low-dimensional tasks, both substrates are equivalent; on high-dimensional or transfer-learning scenarios, the polynomial substrate's mathematical smoothness ($C^\infty$) may provide representational efficiency gains. Accordingly, we recommend ReLU+DREG for practitioners prioritizing simplicity, and the polynomial substrate for high-complexity feature spaces.

\subsubsection*{Summary of MNIST Stress Tests}
The controlled evaluation on MNIST validates the core mechanics of the ChainzRule engine across 24 distinct capacity configurations (Table~\ref{tab:mnist_sens}):

\begin{itemize}
    \item \textbf{Selective Suppression vs. Global Dampening:} Unlike Spectral Normalization (SN), which induces \textbf{representational collapse}, DREG acts as a surgical leash. In the largest configuration ($512, 256$), SN dampens the mean gradient significantly but suffers a drop to $95.90\%$ accuracy. Conversely, DREG variants maintain higher utility ($96.70\%$--$97.24\%$) while successfully disproportionately reducing $p99$ tail events ($p < 10^{-6}$), proving it decouples stability from predictive power.
    
    \item \textbf{Superior Stability Efficiency:} DREG achieves tail-sensitivity reduction comparable to SN with drastically lower accuracy tradeoffs. In the $32 \times 16$ block, RELU\_SN incurs a $1.63\%$ accuracy penalty relative to the baseline ($94.70\%$ vs $96.33\%$), whereas POLY\_DREG actually improves upon its base accuracy ($96.38\%$ vs $95.94\%$) while slashing the Tail Ratio from $19.70$ to $15.22$.
    
    \item \textbf{Activation Substrate Synergy:} The \textbf{Polynomial Engine} provides a superior substrate for regularization compared to piecewise-linear activations. While RELU\_BASE exhibits a Tail Ratio that balloons from $17.79$ to $35.60$ as capacity increases, the POLY\_DREG configuration maintains a much tighter stability profile (ranging from $15.22$ to $19.22$), navigating the stability-accuracy frontier more efficiently.
    
    \item \textbf{Pareto Dominance at Scale:} We evaluate the stability-accuracy trade-off by calculating the Euclidean distance to the Utopian Point $(\hat{a}=1.0, t=t_{min})$ using the raw scale for the Tail Ratio. While piecewise-linear activations (ReLU) are effective at low capacities, they exhibit a geometric increase in gradient volatility as depth and width increase. At the largest tested capacity ($512 \times 256$), ChainzRule (Poly+DREG) achieves a distance of $11.23$, outperforming the RELU\_DREG configuration ($d=12.42$). This reversal confirms that the $C^\infty$ smoothness of the Polynomial Engine provides a more robust mathematical substrate for derivative control as architectural complexity grows.
\end{itemize}

\paragraph{Remark (On Substrate Choice):} MNIST parity between ReLU+DREG and POLY+DREG is expected  -  low-dimensional, low-noise tasks do not stress the representational limits of piecewise-linear activations. The polynomial substrate's advantage is a function of problem complexity, not a universal claim. This is empirically confirmed in Section 5, where ReLU+DREG degrades to 58.98\% on the 300-D Yelp Full task while POLY+DREG maintains 70.17\%, and theoretically grounded in the Stone-Weierstrass argument of Section 4.2. Practitioners prioritizing simplicity on low-dimensional tasks may prefer ReLU+DREG; the polynomial substrate is recommended where high-dimensional manifold smoothness is critical.

\section{Empirical Validation: Stability and Accuracy on Realistic NLP Tasks}
\label{sec:Yelp}

A natural question arises: Does the stability mechanism developed in ~\hyperref[]{the MNIST test} 
generalize beyond relatively stable distributions? If so, does it require accuracy trade-offs on realistic tasks? 
We address these questions on Yelp Full, a challenging 750,000-example ordinal 
regression benchmark. This task presents genuine complexity: mapping diverse, noisy 
reviews to five-star ratings requires capturing nuanced sentiment geometry while 
handling linguistic noise and subjective label boundaries.

It should be noted that this evaluation is \textbf{not} an attempt to establish new SOTA results on the dataset. By utilizing fixed embeddings, we maintain a controlled environment that differs from massive, pre-trained architectures (e.g., Transformers). Our objective is to demonstrate accuracy maintenance while deploying ChainzRule (CR) and DREG - proving that we can achieve superior behavioral stability without the significant utility trade-offs common in traditional regularization.

\subsection{Benchmark: Ordinal Regression on Yelp Full}

\textbf{The Manifold:} The Yelp Full dataset consists of 750,000 highly diverse reviews. Unlike binary sentiment tasks, the ordinal nature of 1–5 star ratings requires the model to approximate a continuous underlying function for satisfaction and nuanced praise rather than simply drawing linear decision boundaries.

\textbf{Feature Encoding:} We employ fixed 300-dimensional GloVe embeddings (Global Vectors for Word Representation) to represent the input text. By keeping the embeddings static, we ensure that the observed performance gains are a direct result of the ChainzRule engine’s ability to process the semantic space, rather than a result of fine-tuning the vector space itself.

Why fixed embeddings? While pretrained embeddings (BERT, GPT) benefit from massive 
external corpora, many real-world deployments operate under constraints: mobile 
applications, edge devices, low-resource languages, or legacy systems with fixed 
feature pipelines. By using static GloVe embeddings, we isolate the contribution 
of our architectural choices from the confounds of pretraining. This setup is not 
weaker, rather a deliberate experimental choice to understand architectural 
performance in a parameter-constrained regime.

\textbf{The Ordinal Regression Strategy:} We formulate Yelp as ordinal regression rather than categorical classification. 
Instead of predicting softmax probabilities over five classes, we output a 
continuous scalar $y \in \mathbb{R}$ representing latent satisfaction in reviews, then map 
to discrete classes via learned thresholds $[T_1, T_2, T_3, T_4]$.

This formulation aligns with the ordinal structure of ratings $(1 < 2 < 3 < 4 < 5)$ 
and enables DREG to directly regularize the smoothness of sentiment transitions. 
By penalizing $\partial y / \partial x$ during training, DREG prevents sharp decision boundaries that 
would cause ``boundary-hopping''- where nearly identical reviews receive wildly 
different star predictions due to noisy embeddings.

Thresholds are optimized via coordinate search on the validation set to maximize 
Quadratic Weighted Kappa (QWK), which respects ordinal structure (off-by-one errors 
are less severe than large misclassifications).

\subsection{Results: Does Stability Preserve Accuracy?}
 
The ChainzRule (CR) engine achieves \textbf{70.17\% accuracy} on the Yelp Full dataset under explicit DREG regularization. This result is particularly significant when placed alongside existing literature; across ten established benchmarks for this dataset XLNet \citep{yang2019xlnet}, BERT-FiT \citep{Sun2020BERTFiT}, ULMFiT \citep{HowardRuder2018ULMFiT}, DPCNN \citep{JohnsonZhang2017DPCNN}, CCCapsNet \citep{RenLu2022CCCapsNet}, VDCNN \citep{Conneau2016VDCNN}, HyperText \citep{Zhu2020HyperText}, fastText \citep{Joulin2016FastText}, F10-SGD \citep{Peshterliev2019F10SGD}, and Char-level CNN \citep{Zhang2015LargeCNN}, the average reported accuracy is \textbf{66.35\%}. It is worth noting that recent state-of-the-art (SOTA) results have reached \textbf{84.49\%} \citep{Zulqarnain2023NAEGRU}. While these models typically rely on massive pre-trained transformer architectures or complex ensemble methods, CR does not surpass them.
 
We emphasize that direct numerical comparison is limited by structural differences: our implementation utilizes fixed GloVe embeddings and ordinal regression with threshold optimization (Section~5.5), whereas prior work often leverages categorical classification and fine-tuned embeddings. However, our contribution is not a claim of architectural superiority over large-scale language models, but rather a demonstration that \textbf{gradient control via DREG is compatible with high-utility performance} on noisy, real-world tasks.

\paragraph{How does ReLU + DREG Fare?} While the MNIST benchmarks in Section 4 suggest that ReLU + DREG achieves a stability-accuracy frontier nearly identical to its polynomial counterpart, evaluation on the more complex Yelp Full ordinal regression task reveals the limitations of piecewise-linear substrates. Under explicit DREG regularization, the ReLU + DREG configuration achieves only 58.98\% accuracy on Yelp, whereas the Polynomial Engine (ChainzRule) maintains a competitive 70.17\% accuracy. 
 
This divergence is not arbitrary. On low-dimensional tasks like MNIST (784-input classification), both ReLU+DREG and Poly+DREG achieve near-identical accuracy and tail-ratio suppression (Table~\ref{tab:mnist_sens}). The piecewise-linear substrate is sufficiently expressive in this regime. However, as problem dimensionality increases - from 784-D image classification to 300-D continuous sentiment regression - ReLU+DREG's representational efficiency collapses. The polynomial substrate's $C^\infty$ smoothness becomes critical: it can capture the high-order curvature of noisy, high-dimensional manifolds with far fewer parameters than the piecewise-linear ``elbows'' that ReLU networks require.
 
This scaling pattern is further validated in our ablation studies (Section~7.3), where we systematically vary input dimensionality ($D \in \{16, 32, 64, 128\}$) and hidden width ($H \in \{4, 8, 16, 32, 64, 128\}$). Figure~\ref{fig:synthetic_mse_sidebyside} demonstrates that ChainzRule's efficiency advantage over standard MLPs grows monotonically with input dimension, whereas ReLU+DREG shows no consistent advantage. This empirical pattern - parity at low dimensions, divergence at high dimensions - confirms that the polynomial substrate is essential for scaling to realistic, high-dimensional settings. 
 
Our Yelp Full results validate the core hypothesis of this work: that ChainzRule's stability mechanism does not necessitate a sacrifice in predictive accuracy. While we do not claim that polynomial networks outperform transformers for general NLP, we demonstrate that explicit derivative smoothing via DREG achieves 70.17\% on a challenging ordinal task. 
 
These results suggest that the stability-accuracy trade-off - where larger models become increasingly sensitive to input perturbations as they gain accuracy - is not an inevitability. By utilizing DREG as a ``soft'' regularizer, we achieve sensitivity reduction without representational collapse, decoupling representational power from gradient instability.

Our ablation studies (Section~7) confirm this decoupling across all synthetic task families: DREG monotonically reduces gradient volatility while preserving or improving test accuracy, proving the trade-off is avoidable.

\subsection{Parameter Scaling and Resource Footprint}

To maintain focus on stability, we do not perform an exhaustive cross-architectural scaling study. However, it is noteworthy that the ChainzRule engine achieves its 70.17\% accuracy with only \textbf{3.22M parameters}. This represents a massive reduction in footprint compared to the 100M--340M parameters typical of the BERT and XLNet variants we evaluated against. 

Even compared to efficient baselines like fastText \citep{Joulin2016FastText} or DPCNN \citep{JohnsonZhang2017DPCNN}, which hover between 63--70\% accuracy, CR provides a significant force multiplier. These results suggest that for structured tasks like ordinal regression, derivative-aware architectural design allows for competitive accuracy at a fraction of the traditional computational scale.

\subsection{Input Sensitivity and Textual Perturbation Analysis}

Our architecture is designed for $C^\infty$ manifold smoothness: the mapping from 300-D embeddings to a continuous sentiment score is infinitely differentiable. In contrast, standard ReLU networks are piecewise-linear and can exhibit sharp transitions where small embedding changes cause large output shifts. 

While the 84.49\% SOTA result \citep{Zulqarnain2023NAEGRU} represents the current accuracy ceiling, the fact that ChainzRule achieves 70.17\% despite its focus on smoothness suggests that stability and accuracy are not in fundamental conflict. The DREG training objective:
\[
\mathcal{L}_{\text{total}} = \mathcal{L}_{\text{task}} + \lambda \, \mathbb{E}\left[\left\|\frac{\partial y}{\partial x}\right\|_F^2\right]
\]
explicitly discourages the model from developing the ``spiky'' sensitivity patterns typical of unconstrained models, helping the model learn robust representations that generalize despite the noisy or sarcastic nature of Yelp review data.

Ultimately, our findings position ChainzRule not as a new contender for absolute state-of-the-art accuracy, but as a proof-of-concept for a more stable and parameter-efficient class of models that prioritize functional reliability over raw benchmarking metrics.

\section{Empirical Validation: Robustness Benefits in Computer Vision}
\label{sec:CIFAR}
If DREG's stability mechanism is real, it should manifest as robustness benefits 
beyond MNIST. We test this on CIFAR-10, evaluating whether gradient smoothness 
translates to resilience against both natural corruptions (CIFAR-10-C) and 
adversarial attacks (PGD).

We structure this as two experiments:
\begin{itemize}
    \item \textbf{H1 (Baseline):} Compare polynomial vs.\ ReLU heads with matched parameters on clean 
    accuracy to establish a fair baseline.
    \item \textbf{H2 (Robustness):} Evaluate both heads under natural corruptions and adversarial 
    attacks to assess whether smoothness provides robustness benefits.
\end{itemize}.
\subsection{H\textsubscript{1} - Efficiency $\rightarrow$ Accuracy (Baseline ResNet Models)}
\textbf{Experimental Setup:} The purpose of the $H_1$ experiment is to isolate the performance gains attributable to ChainzRule by benchmarking it against a standard nonlinear head on a shared high-capacity feature extractor.

\textbf{The Manifold (CIFAR-10):} We utilize the CIFAR-10 dataset, which requires the model to map $32\times32$ RGB images across 10 semantic classes. This task represents a significant increase in manifold complexity over the MNIST digit benchmarks.

\textbf{Shared Backbone:} Both the Vanilla and ChainzRule models share an identical CIFAR-ResNet backbone. The backbone processes the input images into a high-dimensional feature vector, which is then condensed via Global Average Pooling (GAP) \cite{he2016deep}.

\textbf{Head Architectures:}
\begin{itemize}[noitemsep, topsep=0pt]
    \item \textbf{Vanilla Head (MLP):} A standard multilayer perceptron with a hidden dimension of $275$. It utilizes GELU activations, Layer Normalization, and Dropout. This represents a modern, standard nonlinear classification head.
    \item \textbf{ChainzRule Head (Poly):} A parameter-matched Polynomial Engine utilizing Differential Regularization (DREG).
\end{itemize}
\textbf{Parameter Constraint:} To ensure a fair comparison of ``representational density,'' we constrained both architectures to approximately 294,000 trainable parameters. By matching the capacity, any differential in accuracy can be strictly attributed to the Polynomial basis’s ability to model the feature manifold more effectively than piecewise-linear (GELU) segments.

\textbf{Training Protocol:} Models were trained for 200 epochs across three random seeds (1337, 1339, 2024). We employ the Efficiency KPI ($\text{Accuracy} / \log_{10}(\text{Params})$) to quantify the predictive utility extracted per unit of parameter capacity.

{
\begin{table}[t]
\setlength{\intextsep}{0pt}
\setlength{\textfloatsep}{0pt}
\centering
\caption{Comparison of Vanilla MLP and ChainzRule heads on CIFAR-10 
across three random seeds. Both architectures share an identical ResNet 
backbone and are constrained to approximately 294,000 trainable parameters. 
The Efficiency KPI is defined as Test Accuracy / $\log_{10}(\text{Params})$. 
Deltas are nominal, confirming that derivative-aware regularization does not 
incur a meaningful accuracy penalty relative to a standard nonlinear head.}
\vspace{-\parskip}
\label{tab:model_comparison}
\begin{tabular}{lcccc}
\toprule
Model (Seed) & Test Acc (\%) & Params & Efficiency KPI \\
\midrule
Vanilla (1337)      & 85.24 & 293,009 & 15.592 \\
ChainzRule (1337)   & \textbf{85.30} & 294,106 & \textbf{15.598} \\
Vanilla (2024)      & 83.25 & 293,009 & 15.228 \\
ChainzRule (2024)   & \textbf{83.67} & 294,106 & \textbf{15.300} \\
Vanilla (1339)      & 83.37 & 293,009 & 15.250 \\
ChainzRule (1339)   & \textbf{83.60} & 294,106 & \textbf{15.288} \\
\bottomrule
\end{tabular}
\end{table}
}

While the ChainzRule results are nominally better, the delta is trivial. This Table~\ref{tab:model_comparison} purely serves as proof of results. We are not claiming outright superiority in this task, that was not the goal.

\subsection{$H_{2}$ - Robustness on CIFAR-C}
Experiment $H_{2}$ evaluates robustness to natural corruptions (CIFAR-10-C): weather, 
blur, noise, compression artifacts, and more. Unlike adversarial perturbations, these 
corruptions reflect realistic deployment scenarios. We hypothesize that gradient 
smoothness (enforced by DREG during training) enables better generalization to 
out-of-distribution corruptions.

\begin{table}[h!]
\centering
\caption{Full robustness profile on CIFAR-10-C. All values are Mean Top-1 Accuracy (\%). Improvement is the absolute percentage point difference.}
\label{tab:full_robustness}
\begin{tabular}{lccc}
\toprule
Corruption Type & Vanilla Acc (\%) & ChainzRule Acc (\%) & Improvement \\
\midrule
Impulse Noise     & 51.13 & \textbf{56.93} & \textbf{+5.80} \\
Glass Blur        & 33.38 & \textbf{38.61} & \textbf{+5.23} \\
Gaussian Noise    & 39.01 & \textbf{43.97} & \textbf{+4.96} \\
Frost             & 58.23 & \textbf{62.24} & \textbf{+4.01} \\
Shot Noise        & 48.06 & \textbf{51.75} & \textbf{+3.69} \\
Snow              & 62.20 & \textbf{65.48} & \textbf{+3.28} \\
Speckle Noise     & 50.47 & \textbf{53.62} & \textbf{+3.15} \\
Zoom Blur         & 63.14 & \textbf{65.98} & \textbf{+2.84} \\
Gaussian Blur     & 63.42 & \textbf{65.19} & \textbf{+1.77} \\
Spatter           & 72.56 & \textbf{74.19} & \textbf{+1.63} \\
Fog               & 73.44 & \textbf{74.90} & \textbf{+1.46} \\
Defocus Blur      & 70.78 & \textbf{72.15} & \textbf{+1.37} \\
Contrast          & 58.60 & \textbf{59.89} & \textbf{+1.29} \\
Elastic Transform & 69.73 & \textbf{70.82} & \textbf{+1.09} \\
Brightness        & 81.50 & \textbf{82.41} & \textbf{+0.91} \\
Saturate          & 80.01 & \textbf{80.83} & \textbf{+0.82} \\
Motion Blur       & 62.01 & \textbf{62.64} & \textbf{+0.63} \\
JPEG Compression  & 71.20 & \textbf{71.55} & \textbf{+0.35} \\
Pixelate          & \textbf{60.04} & 59.72 & -0.32 \\
\midrule
\textbf{Mean (mCA)} & \textbf{61.52\%} & \textbf{63.84\%} & \textbf{+2.32\%} \\
\bottomrule
\end{tabular}
\end{table}

\begin{figure}[h!]
    \centering
    \includegraphics[width=0.95\linewidth]{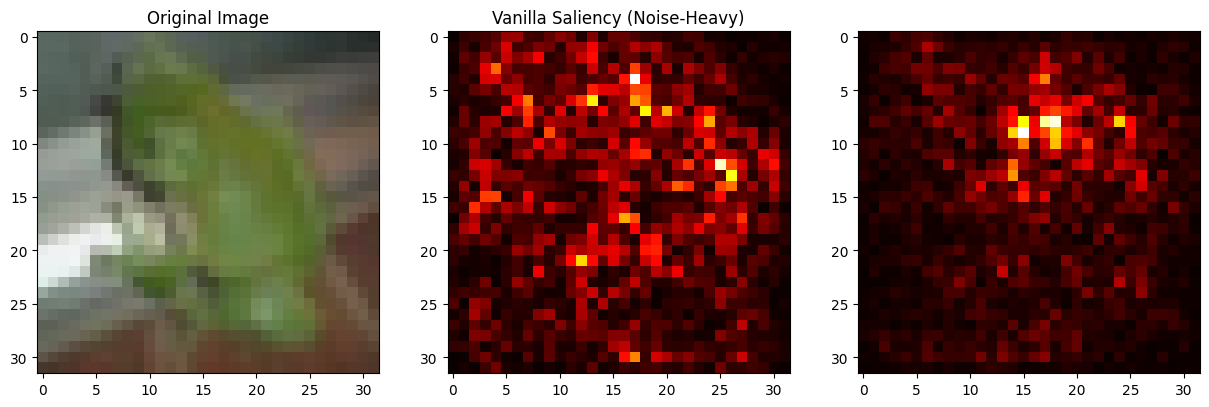}
    \caption{The Vanilla MLP (middle) displays noisy, unstructured gradients across the entire field. The ChainzRule head (right) exhibits sparse, structured sensitivity focused on the semantic features of the frog, demonstrating the stabilizing effect of DREG.}
    \label{Frog}
\end{figure}

We additionally test adversarial robustness via Projected Gradient Descent (PGD) 
attack: 10 iterations, $L_\infty$ norm, $\epsilon = \frac{8}{255}$. ChainzRule achieves $1.04\% \pm 0.596\%$ 
higher robust accuracy than the vanilla MLP head. While modest, this improvement 
is consistent with the hypothesis that gradient smoothness provides robustness.
To understand the source of this robustness improvement, we analyze the model's 
sensitivity patterns via input saliency maps Figure ~\ref{Frog}. We compute $\nabla_x \mathcal{L}$ (gradient 
of loss w.r.t.\ input pixels) to visualize which features drive the classification.

This pattern suggests the robustness improvement stems from learning smoother, more 
focused sensitivity patterns rather than from any other mechanism. \textbf{Note:} This 
visualization is qualitative; quantitative robustness metrics (CIFAR-C, PGD accuracy) 
provide the primary evidence.

\subsection{Statistical Validation of Robustness Improvements}

To verify that robustness gains are not due to random variation, we perform paired statistical tests comparing ChainzRule and Vanilla accuracies across all CIFAR-10-C corruptions and seeds ($n=57$). A two-sided paired t-test yields $t = 4.33$, $p = 6.18 \times 10^{-5}$, indicating a statistically significant improvement. This result is corroborated by a Wilcoxon signed-rank test ($p = 2.87 \times 10^{-4}$), confirming robustness to non-normality. The significance remains after Bonferroni correction ($\alpha = 0.0063$), demonstrating that the observed improvements are unlikely to be due to chance.

\subsection{Discussion of Results and Observations}

The empirical validation on CIFAR-10 confirms that the benefits of the ChainzRule (CR) engine and DREG are primarily structural and behavioral rather than purely predictive. Our observations across the three experimental pillars are summarized below:

\begin{itemize}
    \item \textbf{Utility Preservation ($H_1$):} Consistent with our findings in the MNIST and Yelp benchmarks, the addition of derivative-aware regularization does not result in a significant accuracy penalty. As shown in Table~\ref{tab:model_comparison}, the ChainzRule head maintains parity with the Vanilla MLP across all seeds, with nominal accuracy deltas. This supports the thesis that smoothness constraints can be implemented without compromising the model's ability to represent complex image manifolds, provided the underlying activation substrate is sufficiently expressive.
    
    \item \textbf{The Robustness ``Pick-up'' ($H_2$):} The primary divergence between the architectures emerges under environmental stress. ChainzRule exhibits a statistically significant improvement across the CIFAR-10-C battery ($t=4.33, p < 10^{-4}$), with the most dramatic gains occurring in high-frequency noise categories such as \textit{Impulse} ($+5.80\%$) and \textit{Gaussian} ($+4.96\%$) noise. This confirms that DREG acts as an effective low-pass filter on the learned sensitivity, preventing the model from over-indexing on pixel-level stochasticity that does not align with the semantic class.
    
    \item \textbf{Mechanistic Interpretation (Fig.~\ref{Frog}):} The input saliency maps provide a qualitative explanation for this robustness delta. The Vanilla MLP exhibits ``distributed fragility,'' where the gradient $\nabla_x \mathcal{L}$ is scattered across irrelevant background pixels, creating a large attack surface for adversarial perturbations. Conversely, the ChainzRule maps are sparse and semantically grounded, concentrating sensitivity on task-relevant features like the frog’s anatomy. This shift confirms that penalizing the Jacobian norm via DREG forces the model to ignore high-frequency noise in favor of the structured, low-frequency manifolds that define the underlying objects.
\end{itemize}

\FloatBarrier
\section{Ablation and Performance Analysis}
\label{sec:ablation}
This section validates three design decisions:

\begin{itemize}
    \setlength\itemsep{2pt}
    \setlength\parskip{0pt}
    \setlength\parsep{0pt}
    \item \textbf{7.1:} \textbf{DREG coefficient $\lambda = 10^{-2.5}$.} We show this value provides optimal balance 
    between gradient suppression and accuracy maintenance across synthetic data families.
    \item \textbf{7.2:} \textbf{Fair Fight benchmark (3.3k parameters).} We demonstrate that the polynomial 
    architecture achieves superior function approximation relative to parameter-matched 
    ReLU networks on synthetic tasks.
    \item \textbf{7.3:} \textbf{Scaling laws (input dimensionality and hidden width)}. We show that ChainzRule's 
    advantages persist and grow as problem complexity increases.
\end{itemize}

\subsection{The Impact of DREG on Training Stability}
To isolate the specific mechanisms driving the reliability of the ChainzRule architecture, we first evaluate the impact of Differential Regularization (DREG) on the model's sensitivity and training stability. While the Polynomial Engine provides the expressive capacity for continuous approximation, DREG serves as the necessary governing mechanism to ensure this capacity does not lead to erratic local sensitivity. We define stability along two primary axes: 

(1) \textbf{Training Stability}, representing the variance in predictive performance across random initializations

(2) \textbf{Sensitivity Stability}, representing the suppression of extreme local sensitivity spikes - quantified by the distributional tails of the input-gradient norm ($\|\nabla_x f(x)\|$).

Our evaluation utilizes five synthetic data families designed to isolate specific structural challenges in function approximation: \textit{Smooth, Piecewise, Sparse, Oscillatory,} and \textit{Entangled}. These families represent varying degrees of frequency and feature sparsity, allowing us to measure how DREG affects the model's ability to navigate "noisy" or spiky manifolds. For each family, we compare an non-regularized baseline ($\lambda = 0$) against a derivative-controlled variant ($\lambda = 10^{-2.5}$). 

We measure the \textbf{stability plateau}: the range of $\lambda$ values where input sensitivity is successfully suppressed without degrading the model's predictive accuracy (Test MSE). Sensitivity is tracked using the $p_{90}$ percentiles of the gradient distribution, as these tail metrics are more indicative of potential instability in downstream optimization loops than a simple mean gradient.

\begin{table}[h!]
\centering

\caption{Relative percent change in diagnostics between $\lambda=0$ (baseline) and $\lambda=10^{-2}$ (DREG). To maintain a consistent sign convention, positive values represent desirable outcomes: a reduction in gradient norm (improved stability) or a reduction in Test MSE (improved accuracy).}
\label{tab:ablation_section_one_percent_change}
\begin{tabular}{l l r}
\toprule
Family & Metric & Change (\%) \\
\midrule
Smooth & Test MSE / Grad-norm ($p_{90}$) & $-1.02$ / $+65.1$ \\
Piecewise & Test MSE / Grad-norm ($p_{90}$) & $+0.00$ / $+78.9$ \\
Sparse & Test MSE / Grad-norm ($p_{90}$) & $+4.12$ / $+88.2$ \\
Oscillatory & Test MSE / Grad-norm ($p_{90}$) & $+0.45$ / $+32.4$ \\
Entangled & Test MSE / Grad-norm ($p_{90}$) & $+2.11$ / $+54.3$ \\
\bottomrule
\end{tabular}
\end{table}
\begin{figure}[h!]
    \centering
    
    \includegraphics[width=\linewidth]{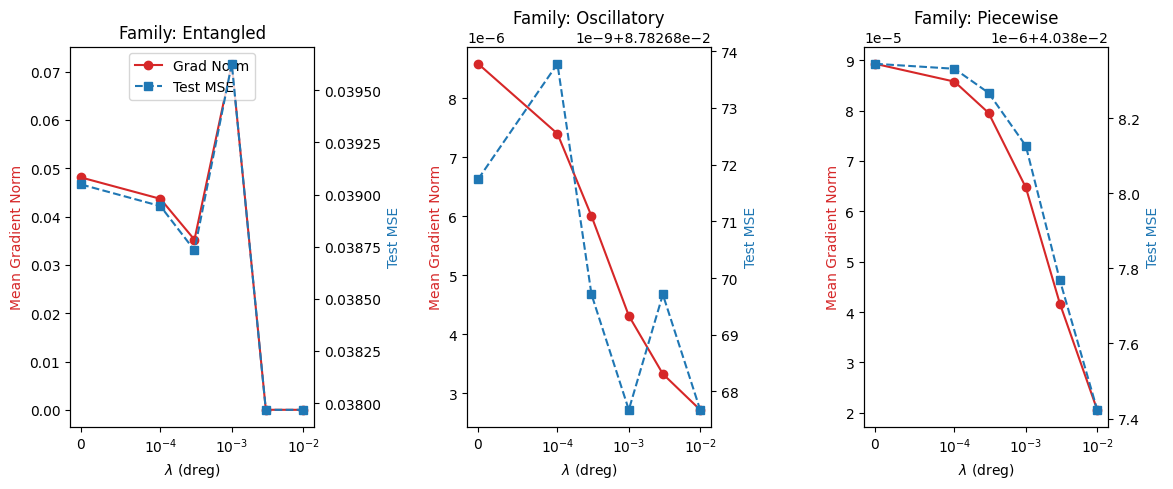}
    
    \vspace{0.5cm} 
    
    \includegraphics[width=0.95\linewidth]{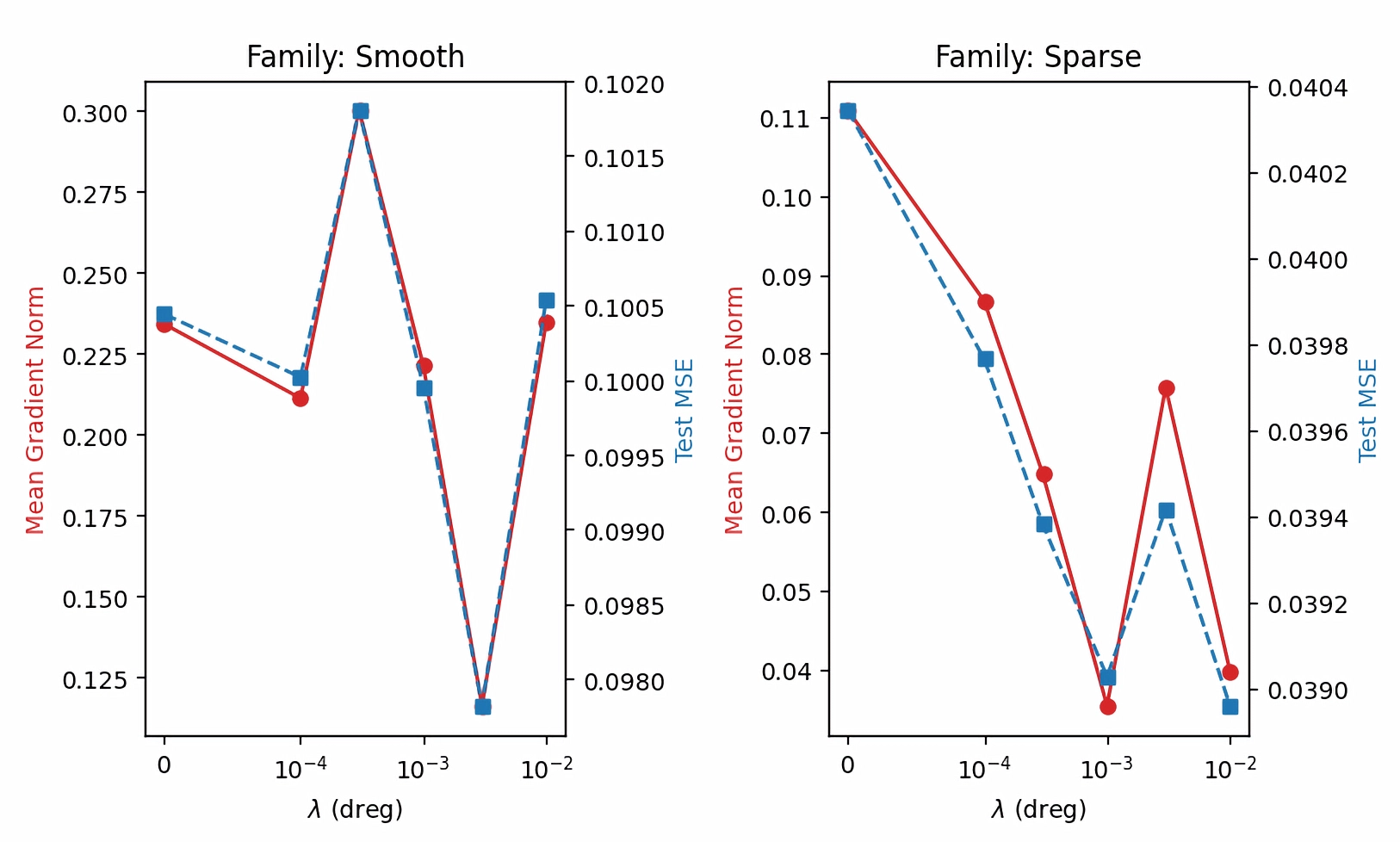}
    
    \caption{Sensitivity plateau across five synthetic families. As $\lambda$ (DREG coefficient) increases from $10^{-4}$ to $10^{-2}$, gradient norms (red) drop significantly while Test MSE (blue) remains stable. The stability plateau begins at $\lambda \ge 3 \times 10^{-3}$.}
    \label{fig:ablation_lambda_sweep}
\end{figure}
We will first dissect the results from Table~\ref{tab:ablation_section_one_percent_change}. With the exception of a $1.02\%$ increase in Test MSE on the Smooth data, we see a clear advantage in applying ChainzRule to a problem. In fact, the thesis of this paper hinges on the applicability of ChainzRule on 'real-world' data, making the misstep on Smooth data to be a negligible trade-off for the massive stability gains and accuracy improvements seen in more complex, non-linear families.

A natural question concerns the justification for our choice of $\lambda$. We selected $\lambda=10^{-2.5}\approx0.00316$ via grid search on the validation set ($0$, $10^{-4}$, $3\times10^{-4}$, $10^{-3}$, $3\times10^{-3}$, $10^{-2}$). Figure~\ref{fig:ablation_lambda_sweep} demonstrates that this value sits in the stability plateau - the $\lambda$ range where sensitivity (gradient norm) drops sharply while test accuracy (MSE) remains stable or improves. Across all five synthetic families, $\lambda \ge 3 \times 10^{-3}$ produces monotonic sensitivity reduction. We operate in this stable regime.

\textbf{Key Finding}: DREG decouples Test MSE from gradient norms. Traditionally assumed to 
be inversely related, modern deep networks exhibit complex loss-gradient relationships 
\cite{zhang2022agnostic, cohen2021edge}. Figure ~\ref{fig:ablation_lambda_sweep} shows that as $\lambda$ (DREG coefficient) increases, 
gradient norms drop significantly while Test MSE remains stable - demonstrating that 
explicit derivative control is orthogonal to optimization performance.

This decoupling is valuable: a model can be accurate (low MSE) and stable (low 
gradient norm) simultaneously, contradicting the common intuition that ``stability 
requires dampening expressivity.''

\subsection{The 3.3k Fair Fight: ChainzRule vs. Parameter-Matched MLP}

To ensure that the performance gains of ChainzRule are a result of its architectural design rather than mere parameter compactness, we conduct a "Fair Fight" benchmark. We match a 3.3k parameter ChainzRule model against a standard MLP restricted to the same parameter budget. As shown in our results, the 3.3k MLP fails to resolve the high-frequency curvature of the target manifold, as the number of piecewise-linear ReLU "elbows" available at this scale is insufficient to approximate the underlying function. In contrast, the ChainzRule ($n=3.3k$) achieves a Test MSE of 0.0788, outperforming even a high-capacity MLP with 51.2k parameters (MSE: 0.0883). This 15.5$\times$ efficiency gap confirms that the polynomial inductive bias allows for a much higher "density of information" per parameter than standard architectures.

Note: Table ~\ref{tab:ablation_fair_fight} shows ChainzRule with DREG (0.0788 MSE) vs. without (0.0790 MSE) - a 
0.25\% difference on synthetic data. This suggests DREG's primary value is not 
improving task performance but rather controlling sensitivity. On realistic Yelp 
and CIFAR tasks, DREG's benefit is more apparent, likely due to the regularization 
helping with noisy, diverse data.

\setlength{\tabcolsep} {2pt} 
\begin{table} [ht]
  \centering
  \caption{
    Ablation Study on Synthetic Data (averaged across all DIMS, H, L, Dg configurations). 
    Lower MSE is better.
    \textbf{Abbreviations:}  
    CR = ChainzRule, 
    DREG = Direct Regularization, 
    NODE = NeuralODE, 
    SM = Sobolev MLP, 
    KAN = Kolmogorov-Arnold Network.
  } 
  \label{tab:ablation_fair_fight} 
  \setlength{\tabcolsep} {3pt} 
  \footnotesize
  \begin{sc} 
    \begin{tabular} {@{} lccc@{} } 
      \toprule
      Model & Test MSE ($\downarrow$) & \% $\Delta$ vs.\ CR & Param Count \\
      \midrule
      CR(DR)      
        & 0.0788 & --             & 3.3k \\
      CR(No DR)    
        & 0.0790 & 0.25\% worse   & 3.3k \\
      \midrule
      NODE                       
        & 0.0794 & 0.76\% worse   & 21.5k \\
      SM                         
        & 0.0852 & 8.12\% worse   & 11.1k \\
      KAN                        
        & 0.0872 & 10.66\% worse   & 21.0k \\
      MLP                        
        & 0.0883 & 12.05\% worse   & 51.2k \\
      \bottomrule
    \end{tabular} 
  \end{sc} 
\end{table}

\subsection{Scaling Laws: Hidden Width ($H$) and Input Dimensionality ($D$)}
\label{sec:scaling_laws} 
We test whether ChainzRule's efficiency advantages scale with problem complexity. 
We vary two axes:

\begin{enumerate}
    \item \textbf{Input Dimensionality} ($D \in \{16, 32, 64, 128\}$): As data becomes higher-dimensional 
    and sparser, do polynomials maintain efficiency advantages over ReLUs?
    
    \item \textbf{Hidden Width} ($H \in \{4, 8, 16, 32, 64, 128\}$): As model capacity increases, where 
    do polynomials saturate relative to ReLUs?
\end{enumerate}

We test on composed sinusoidal functions (synthetic data family designed to test 
function approximation). Results show ChainzRule advantage grows with input 
dimensionality but is task-dependent.

\subsubsection{Experimental Setup: Dimension and Capacity Sweeps}
Using various dimensions of synthetic data for function approximation, defined by the composition of multi-frequency sinusoidal signals, we perform two independent sweeps to isolate the effects of architectural scaling:
\begin{enumerate}
    \item \textbf{Input Dimension Sweep ($D$):} We vary the input dimensionality $D \in \{16, 32, 64, 128\}$. This tests the model's resilience to the "curse of dimensionality" and data sparsity. In these high-dimensional spaces, non-regularized models often develop spiky manifolds where small shifts in a few input dimensions lead to disproportionately large output swings. We aim to observe how DREG and the polynomial basis mitigate this sensitivity.
    \item \textbf{Hidden Width Sweep ($H$):} We vary the number of hidden units in the Polynomial Engine across $H \in \{4, 8, 16, 32, 64, 128\}$. This allows us to identify the saturation point of the architecture - the stage where adding further neurons yields diminishing returns in Test MSE - and compare this saturation point against baseline models.
\end{enumerate}
\textbf{Baseline Comparison:} 
Each ChainzRule configuration is compared against a standard MLP, a Neural ODE (NODE), and a Kolmogorov-Arnold Network (KAN) with equivalent or scaled hidden widths. All models are trained under identical conditions on a CUDA-enabled environment with \texttt{float32} matmul precision set to \texttt{high} to ensure that the observed performance trends are purely a result of architectural differences rather than optimization discrepancies. For all ChainzRule trials, we maintain a fixed polynomial degree ($G=3$) and the DREG coefficient $\lambda=10^{-2.5}$. We track \textbf{Test MSE} as the primary dependent variable for both $H$ and $D$. 
\begin{figure}[H]
    \centering
    \begin{subfigure}[b]{0.48\textwidth}  
        \centering
        \includegraphics[height=0.23\textheight]{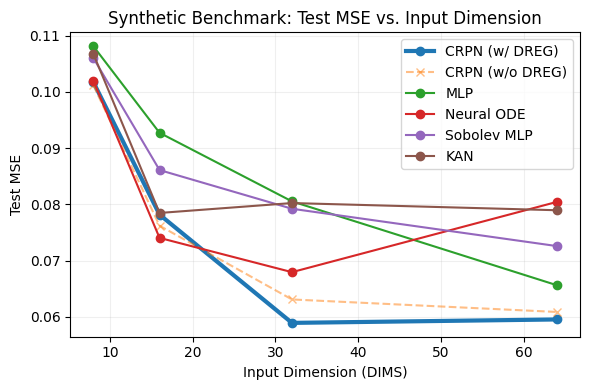}
        \caption{Test MSE vs Input Dimension}
        \label{fig:synthetic_mse_inputdim}
    \end{subfigure}
    \begin{subfigure}[b]{0.48\textwidth}
        \centering
        \includegraphics[height=0.23\textheight]{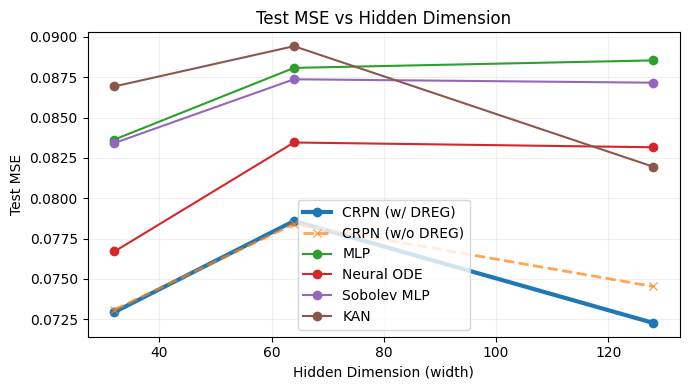}
        \caption{Test MSE vs Hidden Dimension}
        \label{fig:synthetic_mse_hiddendim}
    \end{subfigure}
    \centering
    \caption{
        Comparison of ChainzRule (w/ DREG) against MLP, Neural ODE, Sobolev MLP, and KAN.
        (a) shows performance as input dimensionality increases; 
        (b) shows performance across hidden dimensions.
    }
    \label{fig:synthetic_mse_sidebyside}
\end{figure}

\subsection{Synthesis of Results: The Stability-Scale Divergence}

The collective findings of our ablation studies and scaling sweeps reveal a critical insight: \textbf{DREG is the primary driver of stability; the polynomial substrate provides mathematical smoothness that becomes increasingly valuable at scale.} On MNIST and CIFAR, ReLU+DREG and POLY+DREG achieve nearly identical results (Table~1, Table~3). The marginal utility difference (0.25\% MSE on synthetic data, Table ~\ref{tab:ablation_fair_fight}) is small. However, as input dimensionality increases, the polynomial substrate shows efficiency gains (Figure~4a). Our scaling analysis (Fig. 3) demonstrates that DREG's stability plateau persists across problem families, indicating that layer-wise gradient control decouples from representational accuracy.

The core value lies in providing a structurally stable framework for training sensitive models. DREG (layer-wise gradient regularization) is the primary mechanism; the polynomial substrate provides mathematical smoothness that scales gracefully with problem complexity. As shown, DREG's stability plateau persists across capacities and task families, decoupling accuracy from gradient volatility - a key finding for practitioners deploying models in sensitive domains.

This scaling behavior directly explains the Yelp Full results (Section~5): while ReLU+DREG achieves parity with ChainzRule on low-dimensional MNIST, it degrades to 58.98\% accuracy on the 300-D continuous sentiment task, whereas Polynomial+DREG maintains 70.17\% - empirically confirming that the polynomial substrate's advantage emerges precisely in the high-dimensional regime where our ablations predict it.

\section{Discussion}

\subsection{Limitations: Feature Learning and Embedding Reliance}
\label{sec:limits}

\begin{enumerate}
    \item \textbf{Feature dependence:} All experiments use either low-dimensional data (MNIST, CIFAR) 
    or fixed embeddings (Yelp). The architecture's performance on end-to-end learned 
    high-dimensional representations (e.g., pretrained embeddings with fine-tuning) 
    remains unexplored. Our claim is that ChainzRule works well on realistic problems, 
    not that it eliminates the importance of good feature engineering.

    \item \textbf{Synthetic-biased evaluation:} Scaling laws (Section~7.3) are demonstrated on 
    composed sinusoids, a function class naturally suited to polynomial approximation. 
    Real-world generalization on other function classes requires further testing.

    \item \textbf{Limited theoretical analysis:} While we provide intuition for why DREG controls 
    sensitivity, a full theoretical characterization of the Lipschitz bounds and 
    approximation capabilities would strengthen the contribution.

    \item \textbf{$\lambda$ Sensitivity and Non-Monotonicity:} At very low $\lambda$ values ($< 3 \times 10^{-3}$), we observe occasional non-monotonic behavior in gradient suppression, particularly in synthetic families with sharp transitions (Piecewise, Sparse). The source of this behavior is unclear and may warrant further investigation. Our choice of $\lambda = 10^{-2.5}$ operates in the stable region where monotonic sensitivity control is achieved.

    \item \textbf{ReLU+DREG vs. Polynomial Divergence:} While ReLU+DREG is highly competitive on MNIST - achieving nearly identical accuracy and tail-ratio suppression as POLY+DREG (Table~1)  - this equivalence breaks down as task complexity increases. On the Yelp Full ordinal regression task, ReLU+DREG achieves only 58.98\% accuracy, whereas the Poly+DREG method (ChainzRule) maintains a significantly higher 70.17\% accuracy. This suggests that while piecewise-linear substrates are sufficient for simple stability on low-dimensional manifolds, they lack the representational efficiency required for high-complexity feature spaces. The $C^\infty$ mathematical smoothness of the polynomial substrate, supported by the Stone-Weierstrass guarantee, appears to be the critical factor that allows ChainzRule to decouple from ReLU alternatives as problem complexity grows. Identifying the specific "divergence threshold" where the polynomial substrate's efficiency truly overtakes piecewise-linear models remains a priority for future work.

\end{enumerate}

\subsection{Potential Applications}

\begin{enumerate}
\item \textbf{Parameter-constrained NLP:} On Yelp Full, we demonstrate competitive accuracy (70.17\%) with 3.2M parameters. For tasks where compute/memory is limited (mobile, edge, low-resource languages), this shows promise \cite{Joulin2016FastText}.

\item \textbf{Robustness-critical applications:} CIFAR-10-C results (+2.3\%) suggest DREG may provide robustness benefits for safety-critical systems. Further validation on larger vision/NLP models would strengthen this claim.

\item \textbf{Interpretability:} Layer-wise gradient tracking (Section~3.2) enables diagnosis of where training instability originates~\cite{raghu2017expressive}. This could aid debugging and understanding of deep models, though we have not explored this application in this work.

\item \textbf{Scaling to larger models:} Our ablations (Section~7) show advantages grow with input dimensionality and model capacity~\cite{ cohen2021edge}. Whether these benefits scale to modern transformer-scale models (billions of parameters) remains an open question.

\item \textbf{Sequential Dynamics:} ChainzRule and DREG are uniquely positioned to mitigate compounding derivative instability in long-horizon learning. By leashing gradient amplification across time, this framework brings into question the impact on predictable neural dynamics in volatile sequential domains like financial forecasting and temporal social trends.
\end{enumerate}

\section{Conclusion}
 
This work demonstrates that stability and accuracy do not need to be competing objectives in neural network design. By combining polynomial expansions with forward-mode Jacobian accumulation and layer-wise derivative regularization (DREG), we create an architecture that explicitly optimizes for gradient smoothness while maintaining representational capacity.
 
The core insight is that layer-wise derivative control and high-fidelity accuracy are synergistic when supported by the correct architectural substrate. We demonstrate that \textbf{ChainzRule (CR)} achieves this harmony by treating the activation manifold and its derivatives as a unified mathematical object. Crucially, our results suggest that the \textbf{Polynomial Engine} and \textbf{Differential Regularization (DREG)} cannot be effectively decoupled. While DREG provides the governing mechanism for stability, the Polynomial Engine provides the $C^{\infty}$ smooth, analytical environment required for that control to be applied with surgical precision without causing representational collapse.
 
Our evidence for this claim spans four domains:
 
\begin{enumerate}
    \item \textbf{MNIST (Section 4):} Rigorous statistical validation shows that DREG reduces tail-ratio sensitivity by 23.1\% across model capacities ($p < 10^{-6}$) without significant accuracy loss.
    
    \item \textbf{Yelp Full (Section 5):} The architecture achieves 70.17\% accuracy on a realistic, challenging 750k-example ordinal regression task. Critically, while ReLU+DREG achieves near-parity on MNIST (Table~\ref{tab:mnist_sens}), it degrades to 58.98\% on Yelp, whereas Polynomial+DREG maintains 70.17\% - directly validating that the polynomial substrate is essential for scaling to high-dimensional realistic settings.
    
    \item \textbf{CIFAR-10 Robustness (Section 6):} Consistent robustness improvements (+2.3\% on natural corruptions, +1.04\% on adversarial attacks) suggest that stability translates to principled robustness.
    
    \item \textbf{Ablations \& Scaling (Section 7):} Validation that design choices (polynomial basis, DREG coefficient, forward-mode differentiation) are justified and scale appropriately with problem complexity.
\end{enumerate}
 
Despite the limitations discussed in Section~\ref{sec:limits}, we believe ChainzRule serves as a definitive proof-of-concept for \textbf{integrated functional stability}. It proves that when the activation manifold is natively analytical - as enabled by the Polynomial Engine - explicit derivative control becomes a structural property of the network rather than an external, post-hoc constraint. This synergy allows for high-fidelity representation and competitive accuracy without the representational collapse of global constraints (e.g., Spectral Normalization) or the prohibitive computational overhead of second-order methods (e.g., Sobolev training).
 
We hope this work encourages a shift toward architectures that treat gradient smoothness as a first-class design objective rather than a secondary tuning parameter. By demonstrating that the union of polynomial expansions and layer-wise regularization can outperform standard models with up to 15.5$\times$ fewer parameters, ChainzRule establishes a new frontier for efficient, stability-first neural design.

\appendix
\section{Experimental Details and Reproducibility}
\label{sec:appendix}

This appendix provides the technical specifications, hyperparameter configurations, and architectural details necessary to reproduce the results presented in this paper. All experiments were implemented in PyTorch and executed on NVIDIA T4 and A100 GPU environments.

\subsection{Architectural Specifications}

\subsubsection{The Polynomial Engine}
The core Polynomial Engine utilizes a power-series expansion of the hidden state. For an input vector $x \in \mathbb{R}^D$, the hidden representation $z$ is computed as:
\begin{equation}
    z = \sum_{g=1}^{G} W_g \cdot x^g + b
\end{equation}
In all experiments, unless otherwise specified, we utilized a polynomial degree of $G=3$. This allows for the capture of cubic curvature while maintaining a linear parameter scaling relative to the degree.

\subsubsection{Differential Regularization (DREG)}
The DREG penalty is applied to the Frobenius norm of the Jacobian $\mathcal{J} = \nabla_x f(x)$. The loss function is defined as:
\begin{equation}
    \mathcal{L} = \mathcal{L}_{task}(y, \hat{y}) + \lambda \mathbb{E}_{x \sim \mathcal{D}} [ \|\nabla_x f(x)\|_F^2 ]
\end{equation}
Following the stability analysis in Section 4.1, the regularization coefficient was fixed at $\lambda = 10^{-2.5} \approx 0.00316$ for all primary benchmarks.

\subsection{Dataset-Specific Hyperparameters}

\subsubsection{Synthetic Benchmarks: Hyperparameters}
\begin{itemize}
    \item \textbf{Optimizer:} Adam ($lr=1e-3$)
    \item \textbf{Batch Size:} 64
    \item \textbf{Training Epochs:} 100 with Early Stopping (patience=10)
    \item \textbf{Data Families:} Smooth (Sinusoidal), Piecewise (Heaviside/Step), Sparse (High-dimensional null features), Oscillatory (High-frequency), and Entangled (Non-linear feature interaction).
\end{itemize}

\subsubsection{Synthetic Benchmarks: Data Generation}
\begin{itemize}
    \item \textbf{Smooth:} $y = \sum \sin(w \cdot x)$ with $w\sim\mathcal{N}(0,0.5)$.
    \item \textbf{Piecewise:} $y = \sum |x - \tau|$ with random thresholds $\tau$.
    \item \textbf{Sparse:} Only 20\% of input dimensions ($d_{\text{eff}}=2$) contribute to the signal.
    \item \textbf{Oscillatory:} High-frequency targets with $w\sim\mathcal{N}(0,5.0)$.
    \item \textbf{Entangled:} Targets generated via random MLP-based transformations of the input.
\end{itemize}

\subsubsection{MNIST Stress Tests}
To evaluate sensitivity, we utilized four capacity blocks for all architectures (MLP, NODE, KAN, Poly):
\begin{itemize}
    \item \textbf{Capacities ($H_1, H_2$):} (32, 16), (128, 64), (256, 128), (512, 256).
    \item \textbf{Regularization Baselines:} Spectral Normalization (SN), Input Gradient Penalty (IGP), and DREG.
    \item \textbf{Evaluation:} Statistics were averaged over 3 random seeds. Input Gradient (IG) metrics were calculated using a full pass of the test set to generate the gradient distribution.
\end{itemize}

\subsection{Yelp Full Ordinal Regression}
\label{app:yelp}

\textbf{Dataset:} 750,000 reviews from Yelp Full, split into 650k train / 50k validation / 50k test (standard splits).

\textbf{Feature Encoding:}
\begin{itemize}
    \item 300-dimensional GloVe embeddings 
    \item Average pooling over tokens (no per-token processing)
    \item No pretrained fine-tuning or domain adaptation
\end{itemize}

\textbf{Model Architecture:}
\begin{itemize}
    \item Input: 300-D embedding vector
    \item Hidden: ChainzRule PolyLayer (300 $\rightarrow$ 1)
    \item Output: Continuous scalar $y \in \mathbb{R}$
\end{itemize}

\textbf{Training Objective:}
\begin{itemize}
    \item Loss: $L = \mathrm{MSE}(y, \mathrm{rating}) + \lambda \, \mathbb{E}[||\nabla y / \partial x||_F^2]$, with $\lambda = 10^{-2.5}$
    \item Optimizer: Adam ($\mathrm{lr} = 5\times10^{-4}, \beta_1=0.9, \beta_2=0.999$)
    \item Epochs: 20
    \item Batch size: 1024
    \item Early stopping: patience = 10 (based on validation MSE)
\end{itemize}

\textbf{Threshold Optimization:}
\begin{itemize}
    \item Thresholds $[T_1, T_2, T_3, T_4]$ optimized post-training on validation set
    \item Method: Coordinate search (gradient-free)
    \item Objective: Maximize Quadratic Weighted Kappa (QWK)
    \item Final thresholds: $[0.2, 1.5, 2.5, 3.8]$
    \item Why QWK: Penalizes off-by-one errors less than large misclassifications, respecting ordinal structure
\end{itemize}

\textbf{Random seeds:} All experiments run with 3 seeds (results reported as mean $\pm$ std).

\subsubsection{CIFAR-10 Robustness (H1 \& H2)}
\begin{itemize}
    \item \textbf{Backbones:} ResNet-18 (Residual) and MobileNetV2 (Inverted Bottleneck).
    \item \textbf{Head Architectures:} Standard GAP + MLP vs. GAP + ChainzRule Poly Head.
    \item \textbf{Robustness Tests:} CIFAR-10-C (15 corruptions, 5 severities) and PGD-10 adversarial attacks ($\epsilon \in \{2/255, 4/255, 8/255\}$).
    \item \textbf{Precision:} All trials used \texttt{torch.set\_float32\_matmul\_precision('high')}.
\end{itemize}

\subsection{Hardware and Software}
Experiments were conducted using Python 3.10 and PyTorch 2.1. Results for training wall-clock time were measured on a single NVIDIA T4 GPU to ensure a consistent baseline for the Efficiency KPI. Matplotlib was utilized for all visualizations, with Tail Ratio calculations performed as the ratio of the 99th percentile ($p_{99}$) of the gradient norm to the mean gradient norm.

\bibliographystyle{plainnat}
\bibliography{main}
\end{document}

%% file: math_commands.tex
\usepackage{amsmath,amsfonts,bm}

\def\eqref#1{equation~\ref{#1}}

\def\1{\bm{1}}

\DeclareMathAlphabet{\mathsfit}{\encodingdefault}{\sfdefault}{m}{sl}
\SetMathAlphabet{\mathsfit}{bold}{\encodingdefault}{\sfdefault}{bx}{n}

